%% file: top.tex
\documentclass{article}

\usepackage{arxiv}

\usepackage[utf8]{inputenc} 
\usepackage[T1]{fontenc}    
\usepackage{hyperref}       
\usepackage{url}            
\usepackage{booktabs}       
\usepackage{amsfonts}       
\usepackage{nicefrac}       
\usepackage{microtype}      
\usepackage{lipsum}		
\usepackage{graphicx}
\usepackage{doi}

\usepackage{cite}
\usepackage{amsmath,amssymb,amsfonts}
\usepackage{algorithmic}
\usepackage{graphicx,color}
\usepackage{textcomp}
\usepackage{xcolor}
\usepackage{hyperref}

\usepackage{cite}
\usepackage{amsmath,amssymb,amsfonts}
\usepackage{algorithmic}

\usepackage{units}
\usepackage{pgfplots}
\usepackage{graphicx}
\usepackage{tikz}
\newlength\figureheight
\newlength\figurewidth

\usepackage{textcomp}
\usepackage{xcolor}
\usepackage{units}
\usepackage{subcaption}
\graphicspath{ {./fig/} }

\usepackage{booktabs}
\usepackage{multirow}

\usepackage{glossaries}
\usepackage{balance}

\makeglossaries
\newacronym{nn}{DNN}{Deep Neural Network}
\newacronym{cnn}{CNN}{convolutional neural network}
\newacronym{ai}{AI}{artificial intelligence}
\newacronym{mac}{MAC}{multiply-accumulate}
\newacronym{cim}{CIM}{compute-in-memory}
\newacronym{sram}{SRAM}{static random access memory}
\newacronym{lsb}{LSB}{least-significant bit}
\newacronym{msb}{MSB}{most-significant bit}
\newacronym{dsp}{DSP}{digital signal processing}
\newacronym{pdf}{PDF}{probability density function}
\newacronym{zp}{ZP}{zeropoint}
\newacronym{mos}{MOS}{metal-oxide-semiconductor}
\newacronym{ic}{IC}{integrated circuit}
\newacronym{pe}{PE}{processing element}
\newacronym{noc}{NoC}{network-on-chip}
\newacronym{gg}{GG}{generalized Gaussian}
\newacronym{xor}{XOR}{Exclusive-OR}
\newacronym{xnor}{XNOR}{Exclusive-NOR}
\newacronym{sm}{SM}{Sign-Magnitude}
\newacronym{ip}{IP}{intellectual property}
\newacronym{tcm}{TCM}{tightly coupled memorie}
\glsunset{ai}
\glsunset{ip}
\glsunset{xor}
\glsunset{xnor}
\glsunset{mos}
\glsunset{sram}
\glsunset{cnn}
\glsunset{lsb}
\glsunset{msb}
\glsunset{dsp}

\title{Exploiting Neural-Network Statistics \\for Low-Power DNN Inference}

\author{ \href{https://orcid.org/0000-0003-4673-8310}{\includegraphics[scale=0.06]{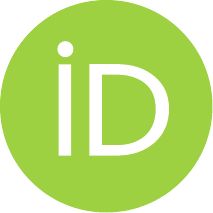}\hspace{1mm}Lennart ~Bamberg} \\
	NXP semiconductors, Hamburg, Germany\\
	\And
	\href{https://orcid.org/0000-0002-6529-4084}{\includegraphics[scale=0.06]{orcid.pdf}\hspace{1mm}Ardalan Najafi} , \href{https://orcid.org/0000-0002-6461-3864}{\includegraphics[scale=0.06]{orcid.pdf}\hspace{1mm} Alberto Garcia-Ortiz} \\
	Integrated Digital Systems, ITEM institute \\
        University of Bremen, Germany\\
}

\renewcommand{\headeright}{ }
\renewcommand{\shorttitle}{Exploiting NN Statistics for Low-Power DNN Inference}

\hypersetup{
pdftitle={Exploiting Neural-Network Statistics for Low-Power DNN Inference},
pdfsubject={cs},
pdfauthor={Lennart Bamberg, Ardalan Najafi, Alberto Garcia-Ortiz},
pdfkeywords={First keyword, Second keyword, More},
}

\begin{document}
\newcommand{\secref}[1]{Sec.~\ref{#1}}
\maketitle

\begin{abstract}
Specialized compute blocks have been developed for efficient \acrshort{nn} execution. However, due to the vast amount of data and parameter movements, the interconnects and on-chip memories form another bottleneck, impairing power and performance. This work addresses this bottleneck by contributing a low-power technique for edge-AI inference engines that combines overhead-free coding with a statistical analysis of the data and parameters of neural networks.
Our approach reduces the interconnect and memory power consumption by up to \unit[80]{\%} for state-of-the-art benchmarks while providing additional power savings for the compute blocks by up to \unit[39]{\%}. These power improvements are achieved with no loss of accuracy and negligible hardware cost.	
\end{abstract}

\keywords{Machine learning \and Edge-AI inference \and Low-power coding \and Low-power digital design \and Neural networks.}

\input{intro}
\input{background}
\input{math}

\input{core}

\input{exp_core}

\input{mac_experimental}
\input{conclusion}

\bibliographystyle{unsrt}
\bibliography{nn_lp}






\end{document}

%% file: intro.tex
\section{Introduction}

The increasing demand for \gls{ai} at the edge has led to the development of various low-cost, yet high-performance \gls{nn} inference engines. However,
due to the relatively large processing requirements and the large memory footprint of modern \glspl{nn} (e.g., the famous \textit{ResNet50} requires \unit[25.6]{M} parameters and \unitfrac[4.1]{GFLOPS}{frame} \cite{he2016deep}),
the power consumption of these inference engines remains a major challenge. This is particularly true for edge-AI deployment in small, battery-operated devices (e.g.,  hearing aids and  IoT nodes). The main contributors to the power consumption of modern edge-AI inference engines are typically the memories and global interconnects used for data storage and exchange. Also, the required parallel \gls{mac} units noticeably impact power consumption.

Therefore, reducing the power consumption of these components is crucial for enabling energy-efficient edge-AI inference. A standard technique in the \textit{TinyML} space (i.e., AI/ML on low-cost/battery-powered devices) is to quantize parameters and activations from 32-bit floats to lower-bit-width integers. However, quantization impacts network accuracy. Thus, to quantize below 8 bits is not possible without harming the accuracy for many applications. Full-integer 8-bit quantization reduces
the power consumption of the memories and interconnects by a factor \unit[4]{$\times$}. The savings for the \gls{mac} units are even higher as integer \gls{mac} blocks are cheaper than the floating-point equivalents~\cite{jacob2018quantization}. Moreover, the complexity of a \gls{mac} unit scales quadratically with the bit width.
Another technique to reduce power consumption at the application level is to use a lighter-weight \gls{nn} (e.g.,~\cite{howard2017mobilenets}). Again, the drawback is a reduced maximum accuracy.

A further advantage of the "lossy" techniques mentioned above is that they substantially reduce the overall memory footprint. This enhances their suitability for systems with small memory capacities. Further techniques to reduce the memory footprint are weight pruning combined with variable-length entropy encoding. Thereby, we iteratively prune weights with a small magnitude (effectively setting them to 0), followed by incremental training to recover accuracy. The large peak in zero-valued weights is then exploited by the entropy encoding, using a lower bit-width for zeros. Variable-length encoding, however, has a negative impact on performance and area due to the decoding hardware and the varying decoder throughput for a given bus bandwidth. Thus, parameters are typically only compressed for off-chip or Flash memories, keeping them uncompressed in the actual \gls{ai} engines.%

The power issue can also be addressed through \gls{cim} or neuromorphic architectures. However, these techniques are likely years away from commercial mass adoption.
Near-memory computing is already deployed at a larger scale. Hence, local \gls{sram} blocks near the compute engines---called \glspl{tcm}---store the weights and activations during processing for cheap data (re)access.
However, to run larger \glspl{nn} or to run multiple networks in a time-multiplexed fashion, additional memory levels are still required to swap parameters or spill activations.
Furthermore, even though better than more traditional architectures, accessing the relatively large local memories still strongly impairs the overall power consumption.

In this paper, we propose a novel approach to further reduce the power consumption of edge-\acrshort{ai} inference engines by combining lossless and overhead-free low-power coding techniques with a comprehensive analysis of the statistical properties of the activations and parameters in neural networks. In particular, our approach exploits the non-maximized entropy of the activation and parameter streams caused by the non-uniform distribution of the data incl.\ activation and weight sparsity. Our approach is meant to extend existing approaches such as light-weight \glspl{nn}, full-integer quantization, pruning, and near-memory-compute, to further optimize the power consumption. Our coding technique is ''lossless'' which implies that it does not affect the \gls{nn} accuracy. Thus, the technique does not impact the outputs of the neural networks at all. This means that no change to the \gls{nn} architectures, the training, or the quantization step is required.
On top, our proposed technique has no noticeable hardware cost and does not increase the memory footprint (i.e., overhead-free coding).

These properties allow designers and architects to apply the technique to their AI engines ''blindly'' (i.e. without a detailed analysis of any trade-offs). 
To the best of the authors' knowledge, it is the first low-power technique for \gls{nn} engines with the property of being universally applicable without drawbacks on any abstraction level.
Despite the easy application of the proposed technique at negligible silicon cost, it reduces the power consumption of interconnects and on-chip memories by over \unit[80]{\%} for real benchmarks in combination with weight-pruning~\cite{han2015deep}---initially proposed as a pure compression technique. On top, we can achieve additional power savings by over \unit[35]{\%} for the processing elements.

Recent research has also addressed
use of low-power coding techniques to decrease the power consumption
in \glspl{nn} accelerators~\cite{10176467}. In particular, it uses bus-invert
for coding the mantissa of the bfloat16 numbers used in a systolic
array. The resulting reduction in switching activity achieves an
overall power reduction of \unit[9.4]{\%} for \textit{ResNet50}. Although the
motivation is similar to ours, this work focuses only on the data-path
of the accelerator, and disregards the potential savings in the memories and interconnects. In addition to the lower overall power savings,
it focuses on floating point numbers instead of integers and employs a
low-power coding technique that increases the bit width of the signals. This increases the hardware complexity as well as the memory and interconnect-bandwidth requirements.

The rest of this paper is structured as follows. Firstly, the
background is presented in \secref{sec:rw}. Subsequently, the bit-level statistics of the parameter and activation streams in modern \glspl{nn} are analyzed. In  \secref{sec:tech}, our proposed technique is presented. 
Afterward, the gains of the technique are qualitatively and
quantitatively assessed in \secref{sec:res}. Finally, a conclusion is drawn in
\secref{sec:conclusion}.

%% file: background.tex
\section{Background}\label{sec:rw}

\subsection{Full-Integer Dynamic-Range Quantization}
 In the following, we describe the standard way of quantizing a neural net to full-integer operations described in \cite{jacob2018quantization}. 

Starting with the input as the first activation set, \glspl{nn} extract activations out of the existing activations successively (feed-forward structure). 
The number of extraction steps is the layer-count/depth of the network. Each step is symbolized as a network layer in a dataflow graph.
The core operation of the \gls{nn} layers is a \acrshort{mac} series of a set of weights $\mathbf{w}$ with a set of activations, $\mathbf{x}$, followed by the addition of a bias, $b$, and a non-linear activation function, $\sigma{()}$, on the result:
\begin{equation}
    y_i = \sigma{\left(\sum_{j} w_{i,j} x_{i,j} + b\right) }
\end{equation}
Different \gls{nn} layer types (e.g., convolutional vs.\ fully connected) differ in how the weights and the activations are gathered and reused for a set of activations to be generated.

Today's standard way of quantizing a \gls{nn} to full-integer operations is described in detail in \cite{jacob2018quantization}. Thus, it is only briefly summarized in the following. 
Different sets of activations or weights are quantized to \textit{int8} (i.e., $q^{x}_i \in [-127,\,128]$) through a \textit{float} scale factor, $S$, and an \textit{int8} \gls{zp}. Thereby, the quantization has a dynamic range based on the value range of each set, determined during training time. The de-quantization from a quantized integer, $q$, back to the \textit{float} is described as:
\begin{equation}
    x_i = S^{x}(q^{x}_i - \mathit{ZP}^x)   
\end{equation}
Hence, the multiply-accumulate can be expressed as: 
\begin{equation}
    S^xS^w\left(\sum_{j} q^w_{i,j} q^x_{i,j} - q^w_{i,j}\mathit{ZP}^x - q^x_{i,j}\mathit{ZP}^w + \mathit{ZP}^x\mathit{ZP}^w + q^b\right)
\end{equation}
Here, the term $-q^w_{i,j}\mathit{ZP}^x$ is constant post-training and can therefore be calculated at compile time and merged into the 32-bit bias for the accumulation. This cannot be done for the term $q^x_{i,j}\mathit{ZP}^w$ due to varying quantized data $q^x$ at inference time. To avoid the extra compute, weights are thus mean-free quantized  (i.e., $\mathit{ZP}^w=0$; $q^w_i  \in [-127, 127]$). It is empirically found that this does not harm accuracy as weight kernels tend to be mean-free due to the applied weight regularisation during training. 

Using highly non-linear activation functions is not only hardware costly but also reduces accuracy for fully integer quantized inference. Thus, full-integer quantization typically focuses on activation functions that are a mere clipping, as the saturation of quantized values to the (\textit{min},\,\textit{max}) values ($-128$,\,$127$) makes this a no operation (\textit{NOP}). The most common activation function for quantized inference is $\textit{ReLU} = \max(x,\,0)$, clipping negative values to zero. With  activation functions that are a mere clipping and symmetrically quantized weights, the core operation becomes:
\begin{equation}
    q^y_i = M{\left(\sum_{j} q^w_{i,j} q^x_{i,j} + q'^{b}\right) }\text{,}
\end{equation}
where ${q'^b}$ is the effective bias equal to $\textit{quant32}(\nicefrac{b}{S_xS_y} + \nicefrac{\mathit{ZP}^y}{M}- q^w_{i,j}\mathit{ZP}^x)$. $M$ is the rescale factor to change to the scale of the quantized output $S_y$ (i.e., $M=\nicefrac{S_wS_x}{S_y}$). The multiplication with this ''float'' factor $M\leq 1$ is in pure integer arithmetic a multiplication of the 32-bit accumulator with a 32-bit integer followed by an 8-bit downshift with rounding and saturation to \textit{int8}.

This formula has the same amount of \gls{mac} operations as the original float variant but all operations are on integers only. Moreover, the cost of executing the activation functions is eliminated by focusing on activation functions that are a mere clipping. Therefore, one rescale operation is needed per output; a 32-bit integer multiplication followed by an 8-bit shift and round with output saturation to \textit{int8}. The biases are 32-bit integers while the \glspl{mac} are entirely done on 8-bit integers. Since typically hundreds to thousands of 8-bit activations are weighted and accumulated per output activation, the cost of having the rescale factors and the bias in 32 bits is negligible. 

\subsection{Low-Power Coding} 
Due to the large parasitic capacitances of modern interconnects, data encoding is a promising low-power technique in advanced technology nodes~\cite{Garcia-Ortiz:2017:1546-1998:356}.
 Low-power encoding adopts the bit-level properties of the data to reduce the power consumption of the physical: interconnects (e.g. metal wires or vias) for data transmission; memory for data storage; or logic for data processing. In this work, we only consider lossless and overhead-free low-power encoding such that it can be applied without further analysis of the impact on the memory footprint, bandwidth requirements, or network accuracy.

 The interconnects in today's \glspl{ic} are made up of metal wires and vias. Metal wires and vias entail a physical capacitance that dominates their power consumption. Whenever the logical value transmitted on a metal interconnect toggles, the capacitance is charged or discharged resulting in an energy loss of \nicefrac{$V_\text{dd}^2C$}{$2$}. Thus, reducing the bit-switching activities of the lines yields a proportional reduction in the dynamic interconnect power consumption. For 3D interconnects, also increasing the 1-bit probability on the lines reduces the power consumption due to the \gls{mos} effect~\cite{bamberg2017high}.

In \glspl{nn}, many parameters and activations are multiply-accumulated to produce a new activation.  Moreover, parameters or activations are typically only loaded/pushed at most once from/to off-chip memories to/from fast local on-chip memories near the \acrlongpl{pe}---potentially with (de)compression on the fly. From these on-chip \glspl{tcm}, they are read hundreds to thousands of times. Thus, the dominating factor for power consumption is reading on-chip memory in modern AI/ML architectures.   

\begin{figure}
    \centering
    \includegraphics[scale=0.9]{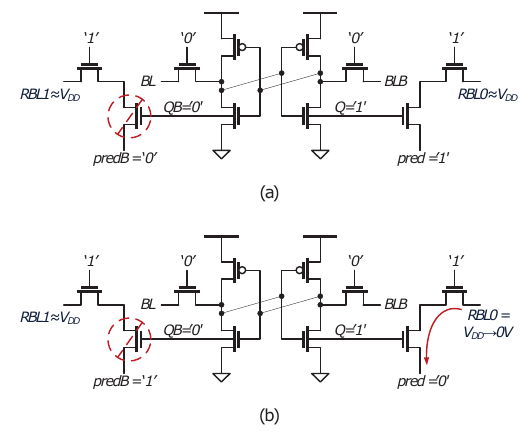}
    \caption{Reading a 0-bit \textit{(a)} and 1-bit \textit{(b)} from a dual-ported 6T \gls{sram} cell. Due to the discharging of the bitline, \textit{(a)} results in a higher power consumption. Figure adopted from~\cite{sinangil2013application}.}
    \label{fig:sram_power}
\end{figure}

 On-chip memories in today's \glspl{ic} are predominantly \glspl{sram}\@. The \gls{sram} read power consumption is dominated by the bitline discharging if the stored value differs from the pre-charge voltage~\cite{sinangil2013application}.
 For single ported \glspl{sram}, low-power coding cannot effectively reduce power consumption due to the complementary read behavior. Always, either the regular or complemented bitline is charged or discharged from $V_\text{pre-charge}$ to zero by the read resulting in a power consumption that is independent of the bits to be read.  
This changes for single-ended reads in dual-ported \glspl{sram}~\cite{sinangil2013application}. Thus, if dual-ported \glspl{sram} are used (e.g., to load parameters for the next operation in parallel to compute) low-power encoding can effectively reduce power consumption via an increased 1-bit probability of the encoded data stored in the memory. This is illustrated in Fig.~\ref{fig:sram_power}.

Lastly, the dynamic power of the logic and sequential elements in the processing elements is again dominated by bit switching of the bits due to internal short-currents when gate inputs switch and the gate input capacitances.
Some flip-flops also show to have a power consumption that depends on the logical bit probabilities.

Leakage is another important power contributor in modern \glspl{ic}, especially in modes of low activity. The leakage power of \gls{sram}, logic, and flip-flops can be optimized again by tuning bit probabilities~\cite{Garcia-Ortiz:2017:1546-1998:356}. Leakage for interconnects is negligible.

In summary, low-power coding should reduce the switching and the bit probabilities. Which metric is more important depends on the components whose power consumption must be reduced.
Generally, \glspl{sram} and global interconnects dominate the power consumption in modern \gls{ai} engines. For the former, the bit probabilities must be optimized, and for the latter the switching probabilities.
The worst case is a completely random switching and 1-bit probability of 0.5 (\unit[50]{\%}). Everything above 0.5 can be transformed to $1-x$ through inverting buffers if the bit-level statistics are understood. This is why our encoding focuses on reducing the 1-bit probabilities even though \glspl{sram} actually needs high 1-bit probabilities for minimal power consumption.

%% file: math.tex
\section{Analysis of the Bit-level Statistics}\label{sec:math}

In this section, we analyze the bit-level statistics of the
streams of standard \textit{int8} (two's complement number representation) quantized weights and activations found in modern Edge-AI inference engines. We moreover,
compare the statistics to the ones for data found in more traditional \gls{dsp} applications.
Data streams sampled from the physical
environment---such as audio, radio signals, etc.--- tend to be Gaussian distributed. 
Consequently, the bit-level
statistics follow a systematic pattern with two characteristic parts:
a group of \glspl{lsb} which are strongly un-correlated and have a switching activity of 0.5, and a group of strongly correlated \glspl{msb} with lower switching activity for temporally correlated signals~\cite{landman1995architectural}. These
bit-level characteristics can be used to decrease the power
consumption during transmission using low-cost low-power codes
\cite{Garcia-Ortiz:2017:1546-1998:356}.
It is worth noting that the bit-level
probability is equal to \unit[50]{\%} for all bits, only the switching
activity changes; and second, that the decrease in switching activity
happens only when the samples are temporally correlated---the typical case in \gls{dsp} signals.

The characteristics of the data streams in \gls{nn} engines are however different.
\begin{figure}[t]
  \centering
  \includegraphics[scale=0.55, trim={0cm 1cm 0cm 2.5cm},clip=True]{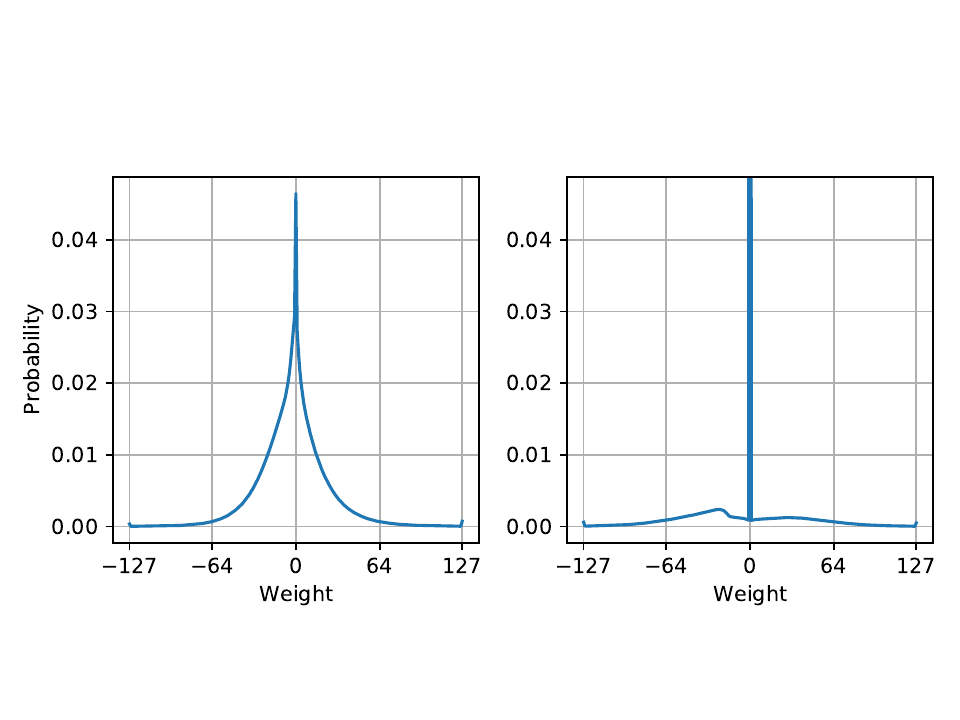}
  \caption{Probability density function of all the weights of an 8-bit-quantized \textit{ResNet50} before (left) and after (right) pruning. After pruning, the
  zero weight contains almost all the probability (\unit[80]{\%}).}
  \label{fig:pdf_weights} 
\end{figure}
The left-hand side of Fig.~\ref{fig:pdf_weights} shows the \gls{pdf} for a stream containing all \unit[25]{M}, per-channel, 8-bit quantized weights of \textit{ResNet50}. 
Analyzing, many \glspl{nn} for this paper showed that light-weight \glspl{nn} such as \textit{MobileNet}, exhibit a typical Gaussian-like distribution with spikes at the edges -127 and 127 due to truncation of the Gaussian tails. However, more heavy \glspl{nn} such as
\textit{ResNets} depict \glspl{pdf} with a larger kurtosis and lower variance, $\sigma_\text{var}$. 
Also, per-tensor instead of per-channel quantization yields a smaller variance, as a single outlier in one channel reduces the values in all other channels.
Bit pruning heavily increases the number of zero values such that it dominates the distribution.

\begin{figure}[t]
  \centering
  \includegraphics[scale=0.55, trim={1cm 1cm 0cm 2.5cm},clip=True]{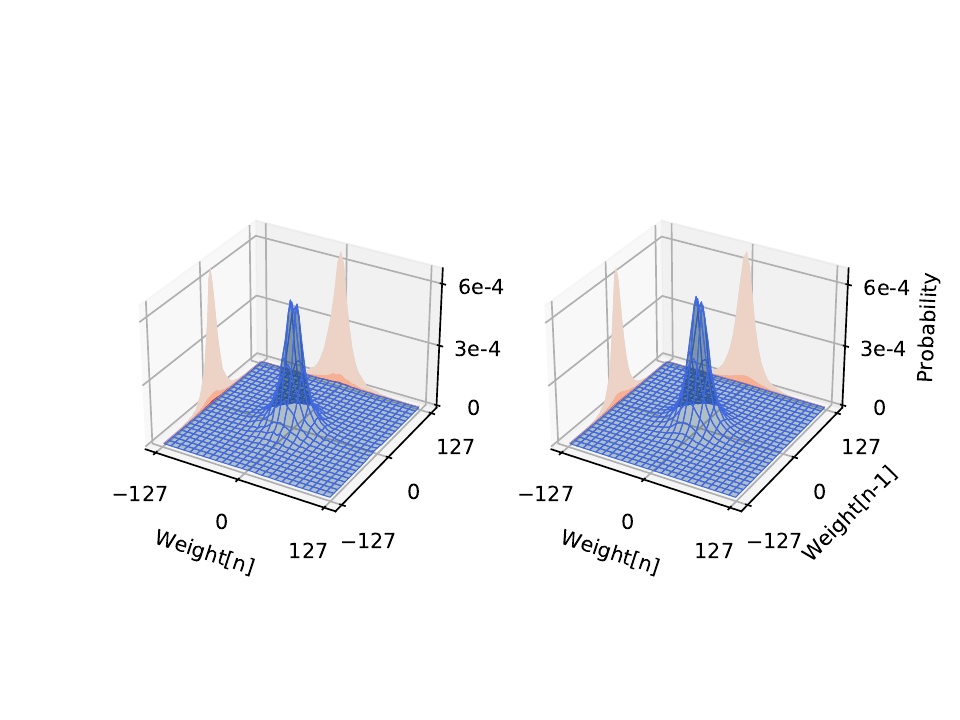}
  \caption{Two-dimensional \gls{pdf} of the weights of a quantized \textit{ResNet50}. Results for 
    two different output layers. The example illustrates the  un-correlation between pattern pairs and the leptokurtic characteristics.}
  \label{fig:pdf_weights_3d} 
\end{figure}

In any case, the switching activity and bit activity are almost exactly 0.5 because the individual weights show to be uncorrelated as shown in Fig.~\ref{fig:pdf_weights_3d}. This yields a particularly high power consumption. The contribution of this paper is to understand the bit-level characteristics of the weights and activations by analytical models and to use these models to derive efficient low-power coding strategies.

For the activation statistics, we constrain our
analysis to the \textit{ReLU} activation function (and its derivatives like \textit{ReLU6}), applying a merge clipping of the values. It is not only the most common activation function for quantized \textit{TinyML} applications; 
it also has great characteristics for low-power processing, as we will show in the remainder of this paper. 
A \textit{ReLU} clips negative values to zero. Thus, a histogram of \textit{ReLU} activations exhibits a large spike at the zeropoint of the quantized number representation. Since a \textit{ReLU} only outputs positive values while 8-bit-quantized activations are in $[-128,\,127]$, the \gls{zp} for \textit{ReLU} activations shows to be always $-128$, transforming the range linearly to a purely positive one, $[0,\,255]$. Since the weighted, accumulated values are normally distributed, after activation, about \unit[50]{\%} of activations are equal to the \gls{zp} $-128$ or 0x80.


\subsection{Modeling Approach}
We empirically found that a good model for the distributions of neural network parameters is the unit-variance \gls{gg}. It is mathematically defined by
\begin{equation}  \label{eq:GG}
GG(x) = \frac{\nu\ \eta(\nu)}{2\ \Gamma(1/\nu)}\ exp\left[- \eta(\nu) \,\|x\|^\nu \right]\text{,}   
\end{equation}
where $\eta$ is a shape parameter that allows parameterizing the
kurtosis of the distribution, given by
$\nicefrac{\Gamma(5/\eta)\ \Gamma(1/\eta)}{\Gamma(3/\eta)^2}-3$.
For a Gaussian distribution, $\eta$ equals 2, and 1 for a Laplacian.


There exists a relation between the entropy rate of a data stream and the maximum power reduction that can be obtained by coding. The relation can be explicitly formulated when
the power depends either only on the transition activity or only on the bit probabilities \cite{775388}.
In this case, the energy consumption is bounded since
the total transition activity of the encoded signal, $T_t$, has the lower bound
\begin{equation}  \label{eq:entropy_bound}
   T_t \geq B \ H_{inv}(\mathcal{H} /B) \text{,}
\end{equation}
 where B is the signal bit-width, $\mathcal{H}$ the pattern entropy, and
 $H_{inv}$ the inverse of the entropy function (i.e. the inverse of $H(p)=-p\,\log_2(p)-(1-p)\log_2(1-p)$)~\cite{775388}.

Using conceptually the same proof as in \cite{775388}, one can show that the total number of expected 0-bits and 1-bits per symbol, $\mathit{PR}_{\{1|0\}}$ have the same lower bound:
\begin{equation}  \label{eq:entropy_bound2}
   \mathit{PR}_{\{1|0\}} \geq B \ H_\text{inv}(\mathcal{H} /B) \text{.}
\end{equation}

%% file: core.tex
\section{Proposed Technique}\label{sec:tech}
As outlined in the previous section, the statistics of the weights and activations show a large potential to decrease power consumption by an appropriate low-power coding.
In this section, we discuss our proposed approach to achieve this reduction. 
First, we give an overview of the type of systems we are
targeting; then, we describe the proposed techniques for encoding the weights and activations; and finally, we analyze the impacts of the encoding variants on the hardware implementation of the MAC.

\subsection{System Overview}\label{sec:sys-overview}
\input{overview}

\subsection{Low-Power Coding for Weights}
In this work, we only consider the most area-efficient coding approaches, inspired by our existing work on more traditional low-power coding~\cite{Garcia-Ortiz:2017:1546-1998:356}. 
Our contribution is a coding framework that, at ultra-low-cost, can change the signal representation to one with optimized switching characteristics or one with optimized bit probabilities, and back. Thereby, \gls{ic} designers can pick for each component (e.g., \gls{sram} or interconnect) the signal representation that results in the best power consumption.

Two separate encoding schemes are required: first, a probability coding that decreases the 1-bit probabilities of the patterns, and second, an \gls{xor}-decorrelator that maps 1-bits at the input $x$ to transitions at the output $y$; i.e. $y = y_\text{prev} \oplus x$. 
To go back from the minimized-switching representation to a minimized-1-bit representation, an \gls{xor}-correlator is used, mathematically expressed as $y = x_\text{prev} \oplus x$.
If 0 bits are minimized by the probability encoding instead of 1 bits, in decorrelator and correlator, \gls{xor} operations must be swapped with \glspl{xnor} to map zeros to transitions. In both variants, encoder and decoder only need $B=8$ flip-flops and X(N)OR gates and have a  logic depth of 1.  

We study two alternatives for the probability coding for \gls{nn} parameters:
\gls{sm} representation and the \gls{xor}-coding
\cite{Garcia-Ortiz:2017:1546-1998:356}. The later approach consists
of \gls{xor}ing the $B-1$ \glspl{lsb} with the \gls{msb} value, leaving the \gls{msb} untouched, for a minimized 1-bit probability.  An advantage of this ''\gls{xor}-\gls{msb} coding" is that the encoder and decoder circuits are identical. Also, it again has ultra-low cost as the coding circuit is $B-1$ parallel \gls{xor} gates (i.e., logic depth of 1). To achieve a minimized 0-bit probability, again simply \gls{xnor} gates are used instead of \gls{xor} gates.

The rationale behind the \gls{xor}-\gls{msb} technique is to exploit that the upper-most bits in each pattern are typically equal (i.e., high spatial correlation), but with equal probability 0 and 1 (i.e., no temporal correlation). By \gls{xor}ing or \gls{xnor}ing each bit with the \gls{msb}, the upper-most bits but the \gls{msb} become mostly 0 or 1, respectively. This by itself does not only optimize the bit probabilities. It already reduces the switching. Thus, the technique optimizes both key metrics at once. Hence,  without additional \gls{xor}-decorrelator coding it can be applied to all kinds of components without additional re-coding steps.

Thus, for the typical case of $B=8$, our proposed coding technique approach only needs 7 \gls{xor}-gates when the \gls{xor}-decorrelator is not used. With the decorrelator-coding, an additional 8 \gls{xor} gates and flip-flops are needed per re-coding. Note that the data is typically already encoded once at compile time with an \gls{xnor}-\gls{msb} encoding to optimize the power consumption of the memories in which the parameters will be stored before they are first used. 

The second variant, \acrfull{sm} encoding, works for \glspl{nn} weights without causing any coding overhead, despite the redundant representation of 0 (0x80 and 0x00), as weights are only symmetrically quantized to integers in $[-127,\, 127]$. Thus, $-128$ does not occur. Sign-magnitude encoding has higher encoder-decoder costs but is based on the same coding idea as the \gls{xor}-\gls{msb} encoding. Hence, it yields similar optimization in terms of bit probabilities and switching.  Despite the higher en/decoding complexity, \gls{sm} encoding for weights is promising, as, for the actual \gls{mac} execution,
weights either need to be decoded or the hardware adjusted to the new weight representation. If the weights are in sign-magnitude format, the hardware becomes actually simpler by the modification compared to the initial $8b \times 8b$ signed \gls{mac}\@. The multiplication becomes 7-bit unsigned (7 \glspl{lsb} of \gls{sm} weight) by 8-bit unsigned activation, followed by a selective addition or subtraction of the result to the accumulator based on the sign-bit of the weight.
Thus, no on-chip \gls{sm} encoding or decoding of the weights is needed at all while the \gls{mac} hardware is minimized. For this scenario, \gls{sm} weights are proposed.

\subsection{Low-Power Coding for Activations}
For activations, the proposed coding is even simpler. The absence of negative numbers due to the \textit{ReLU} activation already heavily reduces the \gls{msb} switching.
Moreover, all bits but the \gls{msb} have low bit probabilities. This is because \unit[50]{\%} or even more patterns are equal to the \gls{zp} (0x80), which only has one 1-bit at the \gls{msb}\@. Thus, X(N)ORing with the constant \gls{zp} effectively optimizes the bit probabilities. For a \gls{zp} of 0x80, this just requires negating the \gls{msb} to reduce 1 bits and negating the 7 \glspl{lsb} to reduce 0 bits. Thus, the hardware cost for this coding we call \acrshort{xor}-\acrshort{zp} is approximately zero. Effectively, the coding moves the number representation from \textit{int8} to \textit{uint8} with a quantization zero-point of 0. This enables directly processing the \textit{uint8} numbers in a \textit{uint} \gls{mac} block without decoding by adjusting the accumulator bias calculated at compile time.

Like for weights, \gls{xor}-decorrelator coding is used to transform optimized bit probabilities into optimized bit switching.
Again, for $B=8$, the proposed decoding approach just needs a couple of gates and 8 flip-flops if the \gls{xor}-decorrelator is present, while the logic depth is one.


\subsection{Implementation of MAC with Coding}
\input{mac}

%% file: overview.tex


Certainly, the exact low-level micro-architectural details of different AI accelerators vary notably; however, the structure of the memory hierarchy and the interconnect architecture is, in the fundamental aspects, very similar.

From bottom to top: AI architectures are composed of a set of \glspl{pe}, typically including a private \gls{tcm} (separated for parameters and activations). 
Multiple \glspl{pe} are clustered in groups that commonly contain one or more additional levels of shared, unified memory dedicated to the AI engine. The individual \glspl{pe} and the dedicated memories are connected over a massively parallel and high bandwidth interconnect architecture---often a \gls{noc}\@. 
The AI engine is connected to the rest of the system over a system-level interconnect. This interconnect architecture can be the same as for the local interconnect between the \glspl{pe} (typically for \glspl{noc}), or a different one such as an AHB or AXI bus.
For large models, the weights are actually saved outside the chip in DRAM, normally in compressed form, and are read on-demand into the on-chip global memories with on-the-fly decompression. 

\begin{figure}[t]
  \centering
  \includegraphics[scale=0.6]{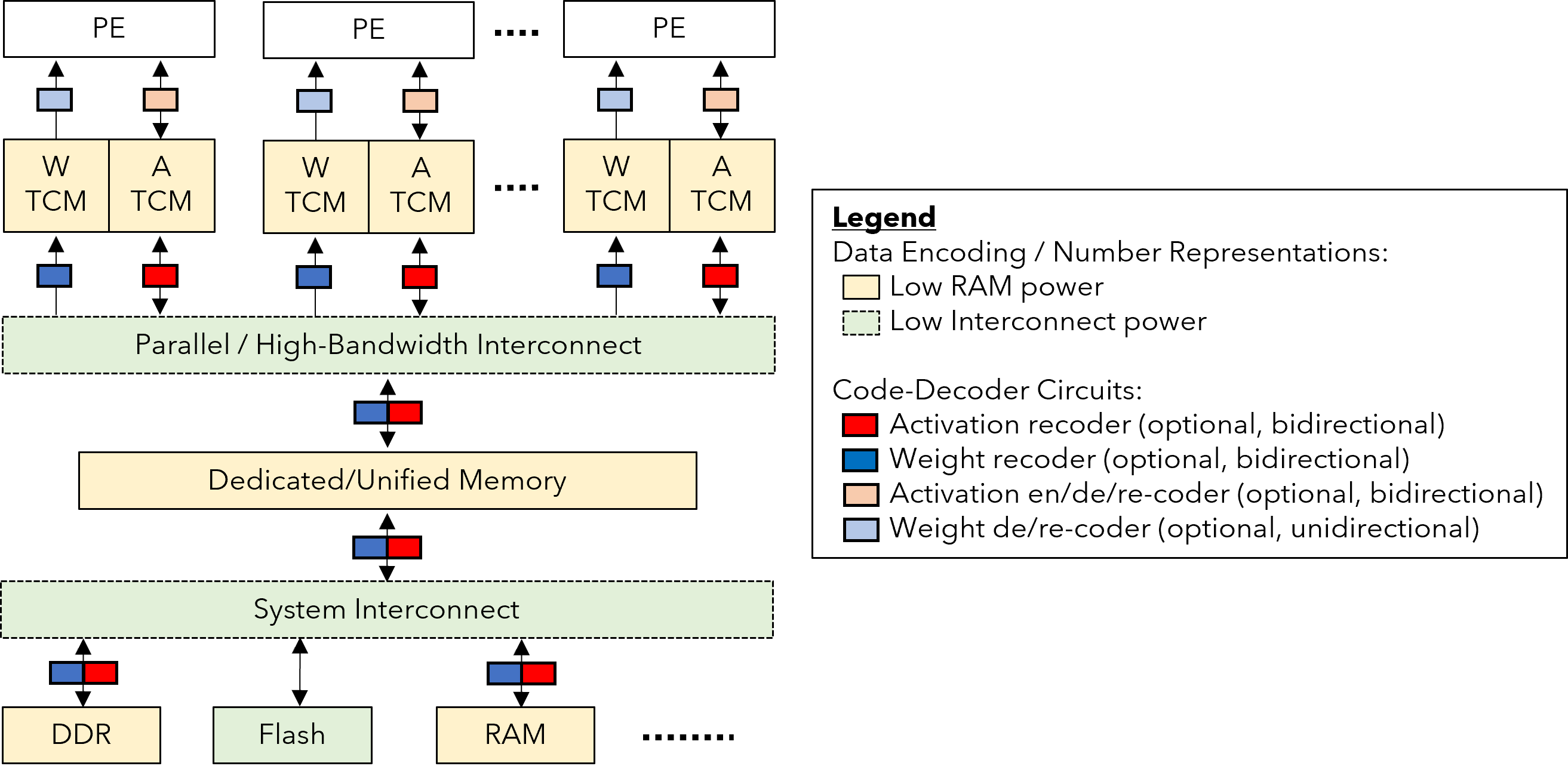}
      \caption{Illustrative system overview with encoding/decoding for weights and activations.}
  \label{fig:sys_overview} 
  
\end{figure}

The two key streams of information in an AI architecture are the weights and the activations.
The interpretation of these two streams requires three levels of abstraction; from the low level of abstraction to the higher one, they are:
\begin{description}
\item[\emph{word}] 
  The physical stream of bits moving through the interconnects, or the values saved into on-chip memories. Can be encoded using approaches such as Bus-Invert, K0, etc. Thus, the bits have to be converted into (interpreted as) a word, which may imply some hardware modules.
 \item[\emph{integer}] 
  Once decoded, the words need to be interpreted as integer numbers. Here again, there are different alternatives, as for example, the use of sign-magnitude, unsigned numbers or the more common two's complement. Note that this interpretation of words into integers determines the actual ALU that is required in the \glspl{pe} (e.g., a standard signed-signed MAC or a sign-magnitude approach).
\item[\emph{real}] 
  Finally, in an AI architecture, the integer numbers are interpreted as a corresponding real number using a linear mapping characterized by a zero point, $ZP$, and a scale, $S$. The rescale process after the integer \gls{mac}-series is responsible for the right interpretation.
\end{description}

  
  

The fundamental idea of this work is to exploit the statistical characteristics of the activation and weights, as well as the degrees of freedom provided by the interpretation as integer and later as real numbers of the words to decrease the power consumption in the interconnect architecture and the memory hierarchy of AI inference engines.

Fig.~\ref{fig:sys_overview} shows an illustrative system overview with the proposed low-power-coding technique in place.  Weights are low-power encoded at compile time and placed in the on-chip memory or DDR with optional additional compression. Activations are dynamic and not known at compile time. They are generated and read by the \glspl{pe}\@. 
Ideally, activations are only stored on the lowest level of the memory hierarchy, made up of the \glspl{tcm} of the \glspl{pe}, but are spilled to higher hierarchies if the capacity of the \gls{tcm} is exceeded. Weights move from the higher memory levels into the \gls{tcm} if the respective layer of the neural network is processed by the \glspl{pe}\@. From the \gls{tcm}, the \glspl{pe} typically access weight multiple times per inference, depending on the weight reuse in the layer.

The goal of our technique is to improve the power consumption across all levels of the interconnect and memory hierarchy. Thus, only at the input of the \glspl{pe} the weights and activations are potentially decoded into the standard \textit{int8} format to perform the \glspl{mac} computations. At the output of the \glspl{pe}, the activations are low-power encoded again.

However, the weights encoded in the parameter \gls{tcm}, can be also read directly by the \glspl{pe} (without decoding) by adapting the architecture of the \glspl{pe} correspondingly, as shown in the experimental results section of this paper.
Similar to the weights, at the input of the \glspl{pe}, the activations can be either decoded or processed in the encoded format.

At the interface between a memory and an interconnect block, the number encoding/representation is potentially changed to one that optimizes the power of the respective physical medium for transmission or storage. Thus, going from a memory to an interconnect the data encoding is changed to one that optimizes the interconnect power consumption and vice versa. 

It is also possible that the same low-power encoding is used for interconnects and memories. Either because a single encoding sufficiently optimizes both power metrics or because the optimization goal is only the power consumption of one structure. In this scenario, there is at most encoder/decoder circuit between each \gls{tcm} and the corresponding \gls{pe}\@.

Anyway, it is important to use low-power codes with negligible hardware implementation costs to ensure that our approach is universally applicable, as it comes in variants where multiple re- or en/de-coder circuits are required in the system. The only exception is weight encoding, which always happens offline at compile time. It thus can exhibit a larger complexity.

In summary, we need low-power codes for memories and interconnects that exploit statistics of activations and weights for maximized power savings. The low-power codes need to be designed to change between the low-power encodings for memories and interconnects at low hardware cost. On top, the \gls{mac} hardware must either directly work on the encoded weights or activations, or a low-cost decoding to the standard \textit{int8} format is required.


%% file: mac.tex

\begin{figure}[t]
  \centering
  \includegraphics[scale=0.5]{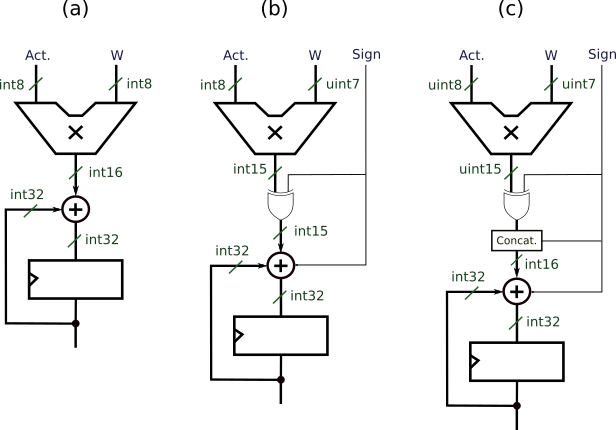}
      \caption{The MAC architectures with inputs: (a) \textit{int8}$-$\textit{int8}, (b) \textit{int8}$-$\textit{uint7}, and (c) \textit{uint8}$-$\textit{uint7}. For the cases where weights are coded with sign-magnitude (\textit{uint7}), the sign bit is used as an additional input to the adder.}
  \label{fig:mac_overview} 
\end{figure}

To demonstrate that the above-mentioned coding schemes result in overall energy savings, we study their effects on the MAC units as well.
Based on the coding schemes discussed earlier, we consider three alternatives for the MAC implementation, as shown in Fig.~\ref{fig:mac_overview}. Note that in the case of XOR-MSB coding, an extra addition is required to obtain the correct result. Since the irregular architecture results in an excessive overhead, the inputs to the \gls{mac} unit should be decoded first for this specific coding. As a result, the corresponding MAC unit would be a normal two's complement (Fig.~\ref{fig:mac_overview}.a), and we do not consider an extra design alternative for XOR-MSB coding.

In Fig.~\ref{fig:mac_overview}.a, the conventional two's complement MAC implementation is shown. This design employs 8-bit signed multiplication. Fig.~\ref{fig:mac_overview}.b illustrates the case where the activations are not coded (\textit{int8}), and the weights are in SM (\textit{uint7}). In this case, the multiplier has one less partial product row, resulting in a smaller architecture. The accumulator input width is reduced by 1 in exchange for the addition of a carry-in and an input XORing (to decode into two's complement). The last alternative shown in Fig.~\ref{fig:mac_overview}.c is the implementation of the MAC unit where the activations are coded with XOR-ZP (\textit{uint8}), while the weights are in SM (\textit{uint7}). In this alternative, an unsigned multiplier is used.

\begin{figure}[b]
  \centering
  \includegraphics[scale=0.43]{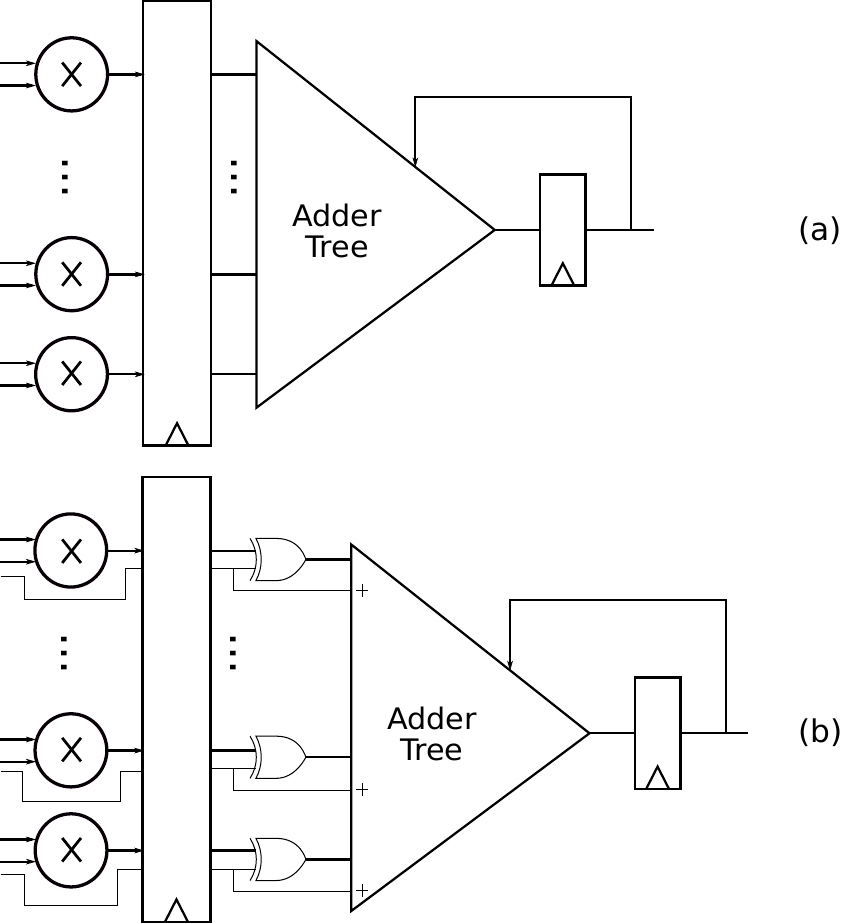}
      \caption{The adder tree-based inner-product units.}
  \label{fig:tree_based_ipu} 
\end{figure}

 To study if the coding ideas mentioned earlier affect the energy consumption of \glspl{pe}, we consider an inner-product unit (IPU) performing eight \gls{mac} operations in parallel. For the baseline case in which no coding is applied, the inner-product unit works with two's complement numbers (see Fig.~\ref{fig:tree_based_ipu}.a). When the weights are in SM, the sign bits are used to negate the outputs of the multiplier by means of parallel XORgates (cf. Fig.~\ref{fig:tree_based_ipu}.b). The additions of the sign bits are integrated into the adder tree so that they do not introduce an overhead. As a result, for the applied codings, the overhead is the XOR gates (8$\times$15 gates for our IPU) and the saving is one partial-product addition for each of the 8 multipliers.

%% file: exp_core.tex
\section{Experimental Results}\label{sec:res}
The experimental results section is divided into three
parts: a qualitative analysis, a quantitative assessment, and the effects of the codings on the energy consumption of the MAC unit.
For a fair, unbiased experimental setup, throughout this work, we analyze only \glspl{nn} that were professionally trained by others. In detail, we use \acrshort{cnn} trained for \textit{ImageNet1k} classification provided by the application module of \textit{Keras} in \textit{TF2.11}. 
This ensures maximized reproducibility of the experimental results without any neural-network IP issues.
With the paper, we further publish a \textit{Python} package to test the power gains of our technique for any quantized \textit{TFlite} model: \textit{LINK BLANKED UNTIL PUBLICATION}. This further enhances reproducibility.

We investigate per-channel quantized neural networks. Per-tensor quantization (the less advanced technique) results in smaller standard deviations for the weight distribution. This
enhances the gains of our technique (cf.~\secref{subsec:quant_res}). 
Thereby, we avoid reporting overly optimistic gains for our technique that are only seen in some applications. 

\subsection{Qualitative Coding Analysis} 
Our qualitative analysis is based on the \unit[25]{M} weights of the standard \textit{ResNet50} architecture as a representative example. The same analysis for other \glspl{nn}  shows no conceptual differences. 
Quantitative results for several \glspl{nn} are provided in the next subsection.
In this qualitative analysis, we investigate the bit-level switching and bit probabilities of the activation and weight data. The order in which the data is processed, read, or transmitted is irrelevant to this analysis, as the values tend to be uncorrelated. Hence, without loss of generality, we can investigate a single random shuffle of the activations and parameters in each tensor.

\begin{figure}[t]
  \centering
  \includegraphics[scale=0.54, trim={0cm 1.5cm 0cm 2.5cm},clip=True]{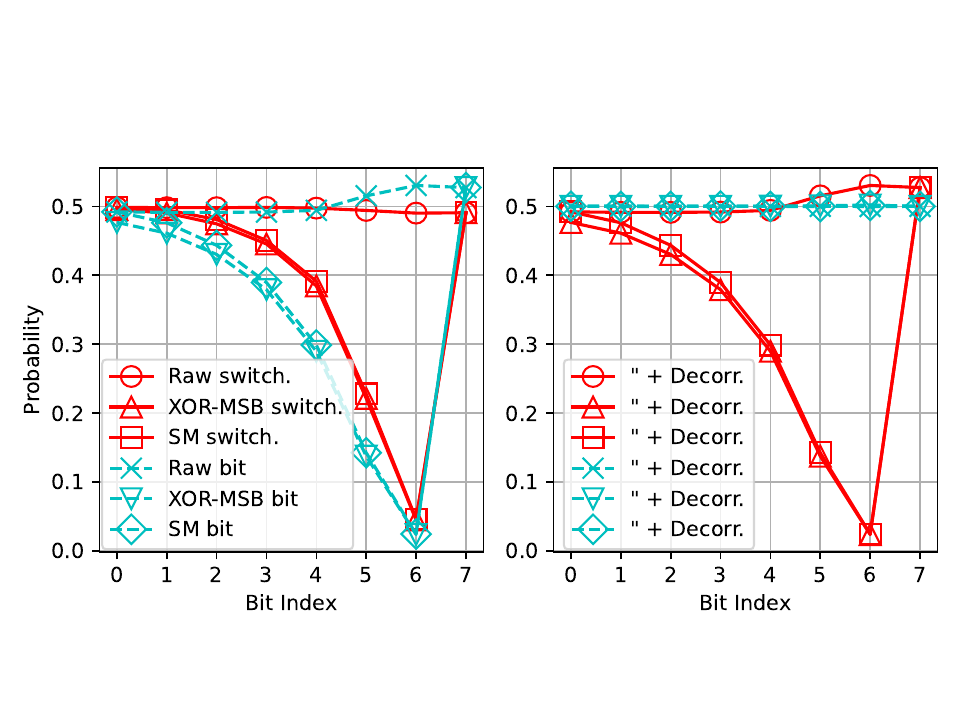}
      \caption{Impact of proposed low-power coding on bit-level statistics for 8-bit quantized \textit{ResNet50} weights (non-pruned): (left) without \gls{xor}-decorrelator; (right) with  \gls{xor}-decorrelator.}
  \label{fig:W_ResNet50} 
\end{figure}

The results for the weights are presented in Fig.~\ref{fig:W_ResNet50}.
The standard raw values result in high power consumption quantities, as both the bit switching as well as the bit probabilities seem to be random (i.e. \nicefrac{\unit[50]{\%}}{bit})---which is the worst case. 
Due to the random bit probabilities an \gls{xor}-decorrelator alone cannot reduce the power consumption, as shown in the raw-switching line of the right panel.

When we apply the \gls{xor}-\gls{msb} coding or the \gls{sm} coding to the weights we observe a significant decrease in the bit probabilities, especially at bit indices 6 to 4. This reduction in the bit probability results in a reduction of switching
activity, as $t_{i} = 2\textit{pr}_i(1-\textit{pr}_i)$ for uncorrelated data. If an XOR-decorrelator is additionally used (right panel),
the switching activity is reduced even further, since 
1 bits are mapped to transitions. The drawback is random bit probabilities again.
Note that there is no improvement in the \gls{msb}, since this bit is not modified by the probability coding schemes.



\begin{figure}[t!]
  \centering
    \includegraphics[scale=0.54, trim={0cm 1.5cm 0cm 2.5cm},clip=True]{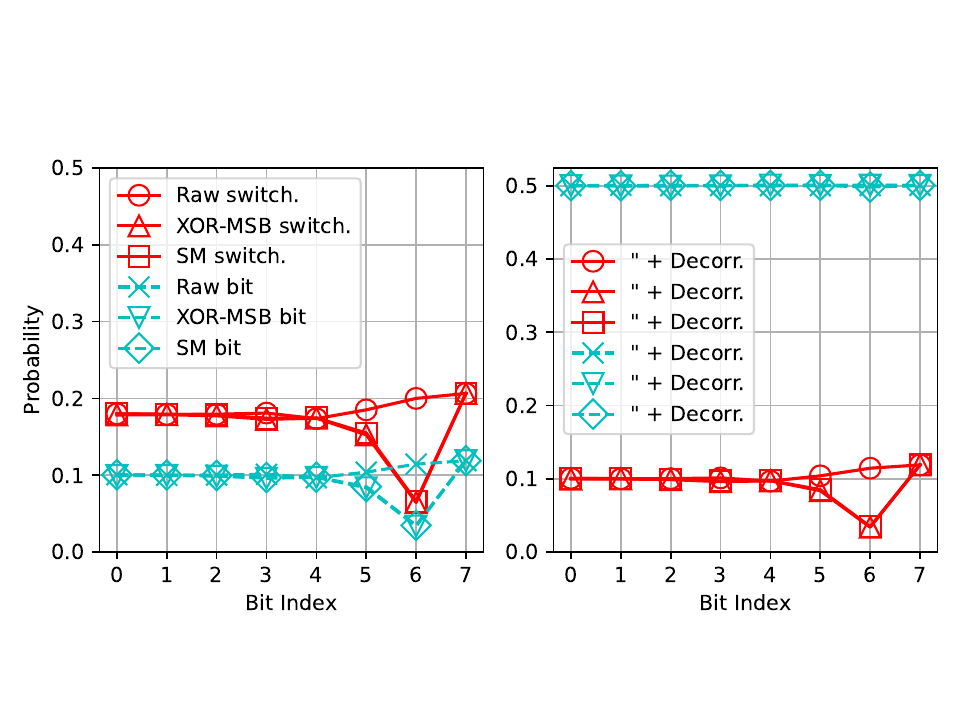}
      \caption{Impact of proposed low-power coding on bit-level statistics for 8-bit quantized \textit{ResNet50} weights that are \unit[80]{\%} pruned: (left) without \gls{xor}-decorrelator; (right) with  \gls{xor}-decorrelator.}
  \label{fig:W_RN-Imagenet-prunned} 
\end{figure}

After the analysis for standard \glspl{nn}, we focus on the effect of weight pruning. The results for the \textit{ResNet50} architecture after pruning \unit[80]{\%} of the weights using the standard \textit{TFLite} pruning methodology are shown in Fig.~\ref{fig:W_RN-Imagenet-prunned}.
The first noticeable effect is that now all the bit probabilities of the unencoded raw data are close to 0.1 instead of 0.5. The reason is the increase of the zero values due to pruning ($ZP=0\text{x}00$ for all weights). This effect reduces the 
switching as well as bit probabilities---and thereby the power consumption--dramatically. After applying our coding approach without an \gls{xor}-decorrelator, 
the switching activity of bits is further reduced by \unit[2]{$\times$}, with a low gain in the \glspl{lsb}\@. When the full coding approach is used with the
\gls{xor}-decorrelator (cf.\ right panel), the switching activity of all bits goes below 0.1. Hence, the total power consumption can be reduced by over \unit[5]{$\times$} for both: components whose power consumption is proportional to the bit switching; as well as components whose power consumption is proportional to the bit probabilities.
It is worth noting that even without further probability coding (e.g. \gls{xor}-\gls{msb} or \gls{sm}), \gls{xor}-decorrelator coding is
very beneficial for pruned weights.


\begin{figure}[t]
  \centering
  \includegraphics[scale=0.54, trim={0cm 1.5cm 0cm 2.5cm},clip=True]{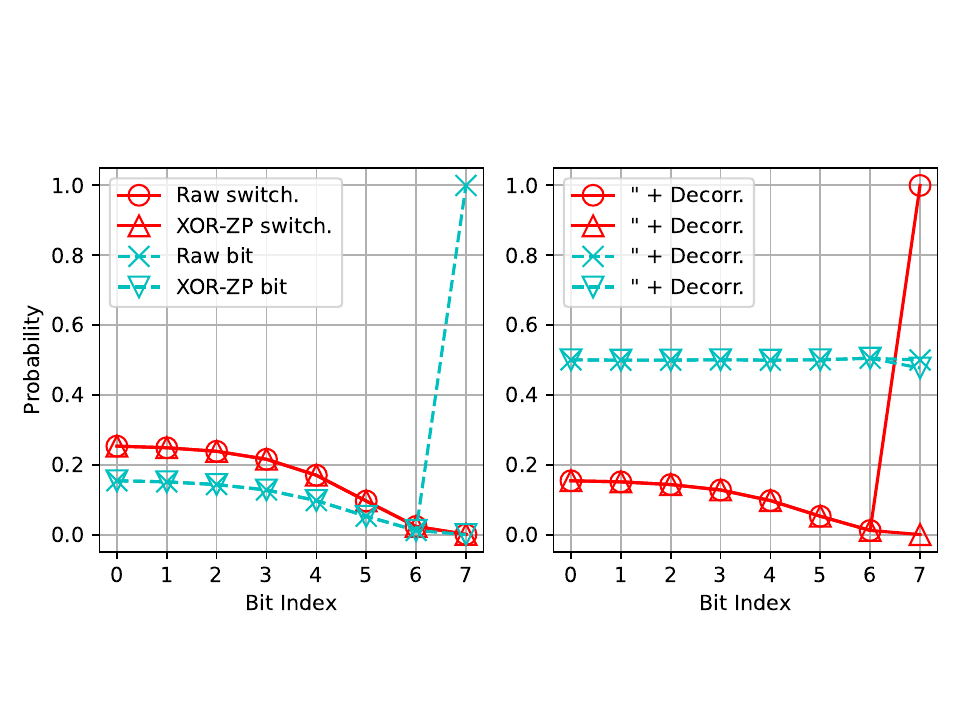}
\caption{Impact of proposed low-power coding on bit-level statistics for all activations within an 8-bit quantized \textit{ResNet50} weights (non-pruned, one input). 
}
  \label{fig:Activation} 
\end{figure}

In the last step of the qualitative analysis, we evaluate our coding for
activations. Again, our analysis for various inferences shows that the gains seem to be relatively independent of the input data, why we represent only data for one randomly selected input. Also, weight pruning shows to have no noticeable impact on the activation distribution why it is not analyzed separately here.
The results for all \unit[9.4]{M} activations of a \textit{ResNet50} inference are shown in Fig.~\ref{fig:Activation}.

The clipping to zero by the \textit{ReLU} activation has a positive effect on the switching and bit probabilities. Compared to zero-mean Gaussian distributed data the switching is reduced by \unit[2]{$\times$}
in the \glspl{lsb} to about \unit[10]{$\times$} in the \gls{msb}\@. The bit probabilities in the \glspl{lsb} are even reduced by \unit[3]{$\times$} for the \gls{lsb} to over  \unit[41]{$\times$} for the second \gls{msb}\@. However, the extremely high bit probability of the \gls{msb} reduces the overall optimization of the bit probabilities to only \unit[2]{$\times$}.
This is fixed by our \gls{xor}-\gls{zp} encoding which here just negates the \gls{msb} resulting in an overall bit-probability and switching optimization compared to random/Gaussian data by over \unit[5]{$\times$} and \unit[10]{$\times$}. Our \gls{xor}-correlator enables to get also \unit[10]{$\times$} switching savings, at no bit probability savings. 
This shows that \textit{ReLU} activations alone already bring the power consumption noticeably down, but our coding can even further push the low-power limits.
 

\subsection{Quantitative Coding Analysis}\label{subsec:quant_res}
In this section, we quantify the switching and bit probability optimization for common \glspl{nn}\@. We analyze two of the most widely used \glspl{cnn} in academia and industry \textit{ResNet50} and \textit{MobileNet}\@. As a cross-validation, we look at MobileNet V1 as well as V2 which differ conceptually but are both light-weight \glspl{nn}\@. Furthermore, we look at a more recent \gls{nn}, \textit{EfficientNet}, to show that the technique is future-proof.
Activations are only investigated for \textit{MobileNetV1} and \textit{ResNet50}, as for others we found no professionally trained but yet publicly available variants with \textit{ReLU} activations. Pruning is as well only reported for these two networks as the other two show the same behavior/gains.

Table~\ref{tab:exp_results} shows the total switching and bit probability reductions obtained with the proposed technique compared to random/Gaussian distributed data. The results for \gls{sm} encoding are similar to the ones for \gls{xor}-\gls{msb} encoding, with a tendency to be slightly worse (always $<$2 percentage points). Thus, \gls{sm} encoding is not listed separately.

\begin{table}[h!]
\caption{Switching and bit probability optimization by the proposed versus the standard technique for various \gls{nn} data.}\label{tab:exp_results}
\centering
\footnotesize
\begin{tabular}{@{}lllll@{}}
\toprule
\textbf{Type}                & \textbf{Encoding}                                    & \textbf{Switching}    & \textbf{Bit Prob.}  \\ \midrule
\multirow{2}{*}{Weights}     &  Standard                                            & \unit[-0.08]{\%}       & \unit[+0.08]{\%}     \\
 \multirow{2}{*}{ResNet50}   & \textbf{\gls{xor}-\gls{msb}}            & \unit[-23.8]{\%}       & \unit[-31.9]{\%}     \\
                             &  \textbf{\gls{xor}-\gls{msb} \& Decorr.} & \unit[-31.9]{\%}       & \unit[+0.03]{\%}     \\ \midrule
\multirow{2}{*}{Weights}     &  Standard                                            & \unit[-0.16]{\%}       & \unit[+1.09]{\%}     \\
\multirow{2}{*}{MobileNet}   & \textbf{\gls{xor}-\gls{msb}}            & \unit[-14.5]{\%}       & \unit[-22.6]{\%}     \\
                             &  \textbf{\gls{xor}-\gls{msb} \& Decorr.} & \unit[-22.6]{\%}       & \unit[+0.01]{\%}      \\ \midrule  

\multirow{2}{*}{Weights}        &  Standard                               & \unit[-0.19]{\%}       & \unit[+0.97]{\%}     \\
\multirow{2}{*}{MobileNet V2}   & \textbf{\gls{xor}-\gls{msb}}            & \unit[-10.8]{\%}       & \unit[-18.5]{\%}     \\
                                &  \textbf{\gls{xor}-\gls{msb} \& Decorr.} & \unit[-18.5]{\%}       & \unit[-0.02]{\%}      \\ \midrule

\multirow{2}{*}{Weights}     &  Standard                                            & \unit[-0.29]{\%}       & \unit[+0.98]{\%}     \\
 \multirow{2}{*}{EfficientNet B0}   & \textbf{\gls{xor}-\gls{msb}}            & \unit[-14.7]{\%}       & \unit[-22.7]{\%}     \\
                             &  \textbf{\gls{xor}-\gls{msb} \& Decorr.} & \unit[-22.7]{\%}       & \unit[-0.25]{\%}      \\ \midrule

\multirow{2}{*}{Pruned Weights}     &  Standard                                            & \unit[-62.9]{\%}       & \unit[-79.1]{\%}     \\
\multirow{2}{*}{ResNet50}   & \textbf{\gls{xor}-\gls{msb}}            & \unit[-67.4]{\%}       & \unit[-81.8]{\%}     \\
                             &  \textbf{\gls{xor}-\gls{msb} \& Decorr.} & \unit[-81.8]{\%}       & \unit[-0.02]{\%}      \\ \midrule    

\multirow{2}{*}{Pruned Weights}     &  Standard                                            & \unit[-61.5]{\%}       & \unit[-78.2]{\%}     \\
 \multirow{2}{*}{MobileNet}   & \textbf{\gls{xor}-\gls{msb}}            & \unit[-64.7]{\%}       & \unit[-80.1]{\%}     \\
                             &  \textbf{\gls{xor}-\gls{msb} \& Decorr.} & \unit[-80.1]{\%}       & \unit[-0.15]{\%}     \\ \midrule

\multirow{2}{*}{Activations}     &  Standard                                            & \unit[-68.1]{\%}       & \unit[-56.8]{\%}     \\
 \multirow{2}{*}{ResNet50}       & \textbf{\gls{xor}-\gls{msb}}                         & \unit[-68.1]{\%}       & \unit[-81.8]{\%}     \\
                                 &  \textbf{\gls{xor}-\gls{msb} \& Decorr.}             & \unit[-81.8]{\%}       & \unit[-0.02]{\%}     \\ \midrule
                                 
\multirow{2}{*}{Activations}     &  Standard                               & \unit[-27.8]{\%}       & \unit[-28.9]{\%}     \\
\multirow{2}{*}{MobileNet}       & \textbf{\gls{xor}-\gls{zp}}            & \unit[-27.8]{\%}       & \unit[-50.4]{\%}     \\
                                &  \textbf{\gls{xor}-\gls{zp} \& Decorr.} & \unit[-50.4]{\%}       & \unit[+0.01]{\%}      \\                               
\bottomrule
\end{tabular}
\end{table}



For standard weights whose individual bits appear random, our technique achieves a power reduction of up to \unit[33]{\%}. This is a remarkable reduction considering that the technique here requires no changes to the training, the \gls{nn} itself, and comes at no accuracy or noticeable hardware cost. For the other lighter-weight \glspl{nn}, the standard deviation of the neural network is smaller. This reduces the possible coding gains. Still, we achieve an optimization by \unit[22.6]{\%}, \unit[18.6]{\%}, and \unit[22.7]{\%} in both power-relevant metrics for \textit{MobileNet V1}, \textit{MobileNet V2}, and \textit{EfficientNet B0}, respectively. 

If weight pruning is applied during \gls{nn} training---which in that case entails no extra cost nor an accuracy loss~\cite{han2015deep}---we even achieve power reductions of over \unit[80]{\%}, also for the lighter-weight \gls{nn}\@. 

We stated in Sec.~\ref{sec:math} of this paper that using \textit{ReLU} activations already reduces the power characteristics versus activation functions for which the 8-bit output tends to be normally distributed. The results show that in fact the power can be reduced by \unit[28]{\%} to \unit[68.1]{\%}. Our proposed technique, which again comes at no accuracy nor noticeable hardware cost, enhances the power savings further to \unit[50.4]{\%} to \unit[81.8]{\%}.

%% file: mac_experimental.tex
\subsection{Energy Consumption of MAC}

This section studies the effects of the codings on the \gls{mac} units. The \glspl{mac} shown in Fig.~\ref{fig:tree_based_ipu} are synthesized in a commercial 40-nm technology using different timing constraints, from relaxed to stringent. The power consumption is measured after back-annotated gate-level simulations. As a representative example, we consider the real activations and weights of an \textit{EfficientNet B0} inference. We report the average energy consumption of the MAC units over all convolution layers (including the depthwise layers). 

\begin{figure*}[t]
  \centering
  \includegraphics[scale=0.98]{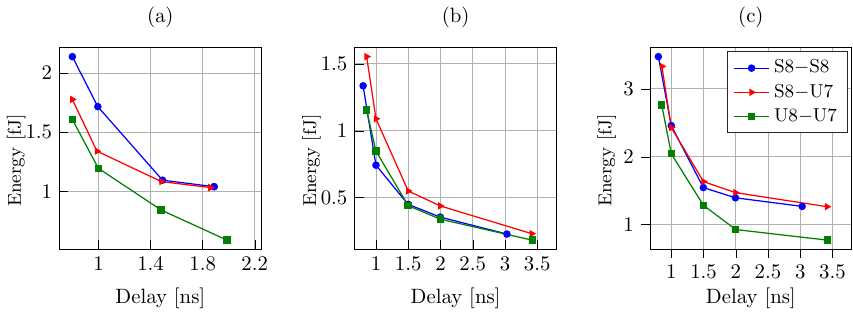}
      \caption{Energy versus delay in a commercial 40-nm technology of: (a) the multipliers, (b) the adder tree, and (c) the total energy of the inner-product unit including multipliers and the adder tree, for different coding schemes}
  \label{fig:compare_mac} 
\end{figure*}

Fig.~\ref{fig:compare_mac} shows the comparison of energy consumption of the multipliers, the adder tree, and the total energy of the \gls{mac} units versus the delay of the units. 
The energy consumptions in the adder trees are marginally different for the studied coding schemes (see Fig.~\ref{fig:compare_mac}.b). In contrast, the coding can result in substantial energy savings in the multipliers (see Fig.~\ref{fig:compare_mac}.a). 
Consequently, as shown in Fig.~\ref{fig:compare_mac}.c, applying low-power coding for memories and interconnects, can even provide additional energy savings in the \glspl{pe} instead of introducing an energy overhead for these components.
 As an instance, when the activations are coded with XOR-ZP (\textit{uint8}), and the weights are in SM (\textit{int7}), up to \unit[39]{\%} energy saving can be achieved in comparison to the case where no coding is applied (activations and weights in \textit{int8}). 
It should be noted that this energy saving is in addition to the aforementioned improvements in memory and communication by the proposed low-power coding framework.

%% file: conclusion.tex
\section{Conclusion}\label{sec:conclusion}
This work presented a thorough analysis of the bit-level statistics of the parameters and activations in full-integer quantized neural networks and an ultra-low-cost coding technique, exploiting the gained insight for power savings. At no noticeable cost, the technique can reduce the power consumption of on-chip memories, interconnects, and logic blocks to up to over \unit[80]{\%}. The technique is effective for any \glspl{nn} with power savings expected in the range of at least \unit[20]{\%} to \unit[30]{\%}. Achieving its full potential requires that the \gls{nn} uses \textit{ReLU} activations and standard weight pruning. On top, the technique can achieve power savings by almost \unit[40]{\%} for the \gls{mac} blocks in the processing elements.
For future work, we will extend this work to further enhance the coding gains for other activations functions that are not a mere clipping such as leaky \textit{ReLU}, or \textit{(H)Swish}. Moreover, we plan to exploit the data statistics also for hardware- and power-efficient compression.